\documentclass[10pt,twocolumn,letterpaper]{article}

\usepackage{iccv}
\usepackage{times}
\usepackage{epsfig}
\usepackage{graphicx}
\usepackage{amsmath}
\usepackage{amssymb}
\usepackage{float}

\usepackage[pagebackref=true,breaklinks=true,colorlinks,bookmarks=false]{hyperref}

\usepackage{overpic}
\usepackage{microtype}
\usepackage{enumitem}
\usepackage{overpic}
\usepackage{placeins}
\usepackage{booktabs}
\usepackage{bbding}
\usepackage{comment}
\usepackage{subfigure}
\usepackage[normalem]{ulem}
\usepackage{multicol}
\usepackage{multirow}
\usepackage{hyperref}
\usepackage{fontawesome}
\usepackage{arydshln}
\usepackage{balance}

\usepackage{hyphenat}
\usepackage[resetlabels,labeled]{multibib}
\newcites{A}{Additional References}

\usepackage{color}
\definecolor{turquoise}{cmyk}{0.65,0,0.1,0.3}
\definecolor{purple}{rgb}{0.65,0,0.65}
\definecolor{dark_green}{rgb}{0, 0.5, 0}
\definecolor{orange}{rgb}{0.8, 0.6, 0.2}
\definecolor{red}{rgb}{0.8, 0.2, 0.2}
\definecolor{darkred}{rgb}{0.6, 0.1, 0.05}
\definecolor{blueish}{rgb}{0.0, 0.3, .6}
\definecolor{light_gray}{rgb}{0.7, 0.7, .7}
\definecolor{pink}{rgb}{1, 0, 1}
\definecolor{greyblue}{rgb}{0.25, 0.25, 1}

\definecolor{blueish}{rgb}{0.0, 0.3, .6}

\renewcommand{\paragraph}[1]{\vspace{.1em}\noindent\textbf{#1}.}

\usepackage{blindtext}

\newcommand{\Fig}[1]{Fig.~\ref{fig:#1}}
\newcommand{\Figure}[1]{Figure~\ref{fig:#1}}

\newcommand{\Table}[1]{Table~\ref{tbl:#1}}
\newcommand{\eq}[1]{\eqref{eq:#1}}

\newcommand{\Section}[1]{Section~\ref{sec:#1}}

\newcommand{\URLicon}{\faGlobe}
\newcommand{\leaderboard}[1]{{\footnotesize \href{#1}{\URLicon}}}

\usepackage{dsfont}
\DeclareMathOperator*{\argmin}{arg\,min}

\newcommand{\loss}[1]{\mathcal{L}_\text{#1}}

\newcommand{\bPhi}{\boldsymbol{\Phi}}
\newcommand{\bx}{\boldsymbol{x}}

\newcommand{\bI}{\boldsymbol{I}}

\newcommand{\IE}{\mathop{\mathbb{E}}}
\newcommand{\calD}{\mathcal{D}}

\newcommand{\given}{~|~}

\newcommand{\encoder}{\mathcal{E}}
\newcommand{\tencoder}{\mathcal{T}_{\mathcal{E}}}
\newcommand{\tdecoder}{\mathcal{T}_{\mathcal{D}}}
\newcommand{\decoder}{\mathcal{D}}
\newcommand{\positional}{\boldsymbol{\Omega}}
\newcommand{\pencode}{\mathcal{P}}

\newcommand{\COTR}{\textbf{COTR}\xspace}
\newcommand{\COTRnobf}{COTR\xspace}
\newcommand{\COTRTnobf}{\COTRnobf+Interp.}
\newcommand{\net}{\mathcal{F}}
\newcommand{\stitchop}{,}

\newcommand{\SupplementaryMaterial}{\textcolor{red}{\textit{supplementary material}}\xspace}

\newcommand{\LiteFlowNet}{LiteFlowNet~\cite{Hui18}~{\tiny CVPR'18}}
\newcommand{\PWCNet}{PWC-Net~\cite{Sun18,Sun19}~{\tiny CVPR'18, TPAMI'19}}
\newcommand{\DGCNet}{DGC-Net~\cite{Melekhov19}~{\tiny WACV'19}}

\newcommand{\GLUNet}{GLU-Net~\cite{Truong20a}~{\tiny CVPR'20}}
\newcommand{\GOCor}{GLU-Net+GOCor~\cite{Truong20b}~{\tiny NeurIPS'20}}

\newcommand{\RAFT}{RAFT~\cite{Teed20}~{\tiny ECCV'20}}
\newcommand{\RAFTP}{RAFT~\cite{Teed20}\xspace}
\newcommand{\COTRT}{\COTR+Interp.}

\definecolor{teaser_red}{RGB}{238,34,13}
\definecolor{teaser_blue}{RGB}{1,161,255}
\definecolor{teaser_green}{RGB}{96,217,54}
\definecolor{teaser_pink}{RGB}{212,25,119}
\definecolor{teaser_orange}{RGB}{242,114,0}
\definecolor{eth3d_green}{RGB}{0,200,0}

\iccvfinalcopy

\ificcvfinal\fi
\begin{document}

\title{\COTR: Correspondence Transformer for Matching Across Images}
\author{
    Wei Jiang$^{1}$, \hspace{3pt}
    Eduard Trulls$^{2}$, \hspace{3pt}
    Jan Hosang$^{2}$, \hspace{3pt}
    Andrea Tagliasacchi$^{2,3}$, \hspace{3pt}
    Kwang Moo Yi$^{1}$
    \\[.2in]
    $^1$University of British Columbia, \hspace{3pt}
    $^2$Google Research, \hspace{3pt}
    $^3$University of Toronto\hspace{3pt}
    
} %

\maketitle

\begin{abstract}
We propose a novel framework for finding correspondences in images
based on a deep neural network that, given two images and a query point in one of them, finds its correspondence in the other.
By doing so, one has the option to query only the points of interest and retrieve \textit{sparse} correspondences, or to query all points in an image and obtain \textit{dense} mappings.
Importantly, in order to capture both local and global priors, and to let our model relate between image regions using the most relevant among said priors, we realize our network using a transformer.
At inference time, we apply our correspondence network by recursively zooming in around the estimates, yielding a multiscale pipeline able to provide highly-accurate correspondences.
Our method significantly outperforms the state of the art on both sparse and dense correspondence problems on multiple datasets and tasks, ranging from wide-baseline stereo to optical flow, \textit{without} any retraining for a specific dataset.
We commit to releasing data, code, and all the tools necessary to train from scratch and ensure reproducibility.
\end{abstract}

\begin{figure}
\centering
\includegraphics[width=\linewidth]{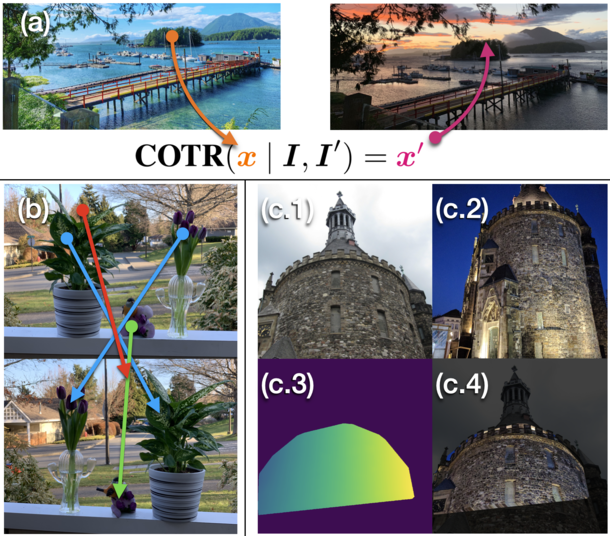}
\caption{
{\bf The Correspondence Transformer -- }
(a) \COTR formulates the correspondence problem as a functional mapping from point {\color{teaser_orange}$\bx$} to point {\color{teaser_pink}$\bx'$}, conditional on two input images $\bI$ and $\bI'$.
(b) \COTR is capable of sparse matching under different motion types, including {\bf {\color{teaser_red}camera motion}}, {\bf {\color{teaser_blue}multi-object motion}}, and {\bf {\color{teaser_green}object-pose changes}}.
(c) \COTR generates a smooth correspondence map for stereo pairs: given (c.1,2)~as input, (c.3)~shows the predicted dense correspondence map (color-coded `x' channel), and (c.4)~warps~(c.2) onto~(c.1) with the predicted correspondences.
} %
\vspace{-3mm}
\label{fig:teaser}
\end{figure}

\section{Introduction}
\label{sec:intro}

Finding correspondences across pairs of images is a fundamental task in computer vision, with applications ranging from camera calibration~\cite{Hartley03,Jin20} to optical flow~\cite{liu2010sift,dosovitskiy2015flownet}, Structure from Motion (SfM)~\cite{schoenberger2016sfm,Jin20}, visual localization~\cite{sattler2012improving, Sarlin19, lynen2020large}, point tracking~\cite{lucas1981iterative, tomasi1991tracking}, and human pose estimation~\cite{newell2016stacked,Guler18}.
Traditionally, two fundamental research directions exist for this problem.
One is to extract sets of \textit{sparse} keypoints from both images and match them in order to minimize an alignment metric~\cite{sift, sattler2012improving, Jin20}.
The other is to interpret correspondence as a \textit{dense} process, where every pixel in the first image maps to a pixel in the second image~\cite{liu2010sift, sun2014quantitative, Zhou17,ummenhofer2017demon}.

The divide between \textit{sparse} and \textit{dense} emerged naturally from the applications they were devised for.
Sparse methods have largely been used to recover \textit{a single global} camera motion, such as in wide-baseline stereo, using geometrical constraints.
They rely on local features~\cite{Lowe04,yi2016lift,Ono18, Detone18} and further prune the putative correspondences formed with them in a separate stage with sampling-based robust matchers~\cite{fischler1981random,barath2019magsac,Chum05}, or their learned counterparts~\cite{yi2018learning,brachmann2019neural,zhang2019oanet,sun2020acne,Sarlin20}.
Dense methods, by contrast, usually model \textit{small} temporal changes, such as optical flow in video sequences, and rely on \emph{local smoothness}~\cite{lucas1981iterative,horn1981determining}.
Exploiting context in this manner allows them to find correspondences at arbitrary locations, including seemingly texture-less areas.

In this work, we present a solution that bridges this divide, a novel network architecture that can express both forms of prior knowledge~--~global \textit{and} local~--~and learn them implicitly from data. 
To achieve this, we leverage the inductive bias that densely connected networks possess in representing smooth functions~\cite{atzmon2020sal,basri2020frequency,rahaman2019spectral}
and use
a transformer~\cite{Vaswani17,Carion20,dosovitskiy2020image} to automatically control the nature of priors and learn how to utilize them through its attention mechanism.
For example, ground-truth optical flow typically does not change smoothly across object boundaries, and simple (attention-agnostic) densely connected networks would have challenges in modelling such a discontinuous correspondence map, whereas a transformer would not.
Moreover, transformers allow encoding the relationship between different locations of the input data, making them a natural fit for correspondence problems.

Specifically, we express the problem of finding correspondences between images $\bI$ and $\bI'$ in functional form, as $x' = \net_\Phi(\bx \given \bI, \bI')$, where $\net_\Phi$ is our neural network architecture, parameterized by $\Phi$, $\bx$ indexes a \textit{query} location in $\bI$, and $\bx'$ indexes its corresponding location in $\bI'$; see~\Figure{teaser}.
Differently from sparse methods, \COTRnobf can match \emph{arbitrary} query points via this functional mapping, predicting only as many matches as desired.
Differently from dense methods, \COTRnobf learns smoothness \emph{implicitly} and can deal with large camera motion effectively.

Our work is the first to apply transformers to obtain accurate correspondences. %
Our main technical contributions are:
\vspace{-1.6em}
\begin{itemize}[leftmargin=*]
\setlength\itemsep{-.3em}
\item we propose a \textit{functional correspondence} architecture that combines the strengths of dense and sparse methods;
\item we show how to apply our method \textit{recursively} at multiple scales during inference in order to compute highly-accurate correspondences;
\item we demonstrate that \COTRnobf achieves \textit{state-of-the-art} performance in both dense and sparse correspondence problems on multiple datasets and tasks, without retraining;
\item we substantiate our design choices and show that the transformer is key to our approach by replacing it with a simpler model, based on a Multi-Layer Perceptron (MLP).
\end{itemize}

\section{Related works}
We review the literature on both sparse and dense matching, as well as works that utilize transformers for vision.

\paragraph{Sparse methods}
Sparse methods generally consist of three stages: keypoint detection, feature description, and feature matching.
Seminal detectors include DoG~\cite{Lowe04} and FAST~\cite{rosten2006machine}.
Popular patch descriptors range from hand-crafted~\cite{Lowe04,calonder2010brief} to learned~\cite{mishchuk2017working,tian2019sosnet,ebel2019beyond} ones.
Learned feature extractors became popular with the introduction of LIFT~\cite{yi2016lift}, with many follow-ups~\cite{Detone18, Ono18, dusmanu2019d2, revaud2019r2d2, bhowmik2020reinforced, Tyszkiewicz20}.
Local features are designed with sparsity in mind, but have also been applied densely in some cases~\cite{tola2009daisy,liu2010sift}.
Learned local features are trained with intermediate metrics, such as descriptor distance or number of matches.

Feature matching is treated as a \textit{separate} stage,
where descriptors are matched, followed by heuristics such as the ratio test, and robust matchers, which are key to deal with high outlier ratios.
The latter are the focus of much research, whether hand-crafted, following RANSAC~\cite{fischler1981random,Chum05,barath2019magsac}, consensus- or motion-based heuristics~\cite{cavalli2020adalam, lin2017code, bian2017gms, ma2019locality}, or learned~\cite{yi2018learning, brachmann2019neural, zhang2019oanet, sun2020acne}.
The current state of the art builds on attentional graph neural networks~\cite{Sarlin20}.
Note that while some of these theoretically allow feature extraction and matching to be trained end to end, this avenue remains largely unexplored.
We show that our method, which does not divide the pipeline into multiple stages and is learned end-to-end, can outperform these sparse methods.

\paragraph{Dense methods}
Dense methods aim to solve optical flow.
This typically implies small displacements, such as the motion between consecutive video frames.
The classical Lucas-Kanade method~\cite{lucas1981iterative} solves for correspondences over local neighbourhoods, while Horn-Schunck~\cite{horn1981determining} imposes global smoothness.
More modern algorithms still rely on these principles, with different algorithmic choices~\cite{sun2010secrets}, or focus on larger displacements~\cite{brox2010large}.
Estimating dense correspondences under large baselines and drastic appearance changes was not explored until methods such as DeMoN~\cite{ummenhofer2017demon} and SfMLearner~\cite{Zhou17} appeared, which recovered both depth and camera motion~--~however, their performance fell somewhat short of sparse methods~\cite{yi2018learning}. 
Neighbourhood Consensus Networks~\cite{rocco2018neighbourhood} explored 4D correlations~--~while powerful, this limits the image size they can tackle.
More recently, DGC-Net~\cite{Melekhov19} applied CNNs in a coarse-to-fine approach, trained on synthetic transformations, GLU-Net~\cite{Truong20a} combined global and local correlation layers in a feature pyramid, and GOCor~\cite{Truong20b} improved the feature correlation layers to disambiguate repeated patterns.
We show that we outperform DGC-Net, GLU-Net and GOCor over multiple datasets, while retaining our ability to query individual points.

\paragraph{Attention mechanisms}
The attention mechanism enables a neural network to focus on part of the input.
Hard attention was pioneered by Spatial Transformers~\cite{jaderberg2015spatial}, which introduced a powerful differentiable sampler, and was later improved in~\cite{jiang2019linearized}.
Soft attention was pioneered by transformers~\cite{Vaswani17}, which has since become the \textit{de-facto} standard in natural language processing~--~its application to vision tasks is still in its early stages.
Recently, DETR~\cite{Carion20} used Transformers for object detection, whereas ViT~\cite{dosovitskiy2020image} applied them to image recognition.
Our method is the first application of transformers to image correspondence problems.~\footnote{A concurrent relevant work for feature-less image matching was proposed shortly after our work became public~\cite{sun2021loftr}.}

\paragraph{Functional methods using deep learning}
While the idea existed already, \eg to generate images~\cite{Stanley07}, using neural networks in functional form has recently gained much traction.
DeepSDF~\cite{Park19} uses deep networks as a function that returns the signed distance field value of a query point. These ideas were recently extended by~\cite{Halimi20} to establish correspondences between incomplete shapes.
While not directly related to image correspondence, this research has shown that functional methods can achieve state-of-the-art performance.

\section{Method}
\label{sec:method}
We first formalize our problem~(\Section{problem}), then detail our architecture~(\Section{architecture}), its recursive use at inference time~(\Section{inference}), and our implementation~(\Section{implementation}).

\subsection{Problem formulation}
\label{sec:problem}
Let $\bx\in [0,1]^2$ be the normalized coordinates of the \emph{query} point in image $\bI$, for which we wish to find the corresponding point, $\bx'{\in}\:[0,1]^2$, in image $\bI'$.
We frame the problem of learning to find correspondences as that of finding the best set of parameters $\bPhi$ for a parametric function $\net_{\bPhi}\left(\bx| \bI, \bI'\right)$ minimizing
\begin{align}
\hspace*{-1mm}
\argmin_{\bPhi} \IE_{\substack{
(\bx,\bx',\bI,\bI')\sim\calD
}} \: \loss{corr} + \loss{cycle},
\label{eq:obj}
\end{align}
\vspace{-1em}
\begin{align}
\loss{corr} &= \left\| \bx' - \net_{\bPhi}\left(\bx \given \bI, \bI'\right)\right\|_2^2 ,
\\
\loss{cycle} &= \left\|\bx - \net_{\bPhi}\left(\net_{\bPhi}\left(\bx \given \bI, \bI'\right) \given \bI, \bI'\right)\right\|_2^2,
\label{eq:cycle}
\end{align}
where $\calD$ is the training dataset of ground correspondences, $\loss{corr}$ measures the correspondence estimation errors, and $\loss{cycle}$ enforces correspondences to be cycle-consistent.

\subsection{Network architecture}
\label{sec:architecture}
\begin{figure}
\centering
\includegraphics[width=\linewidth]{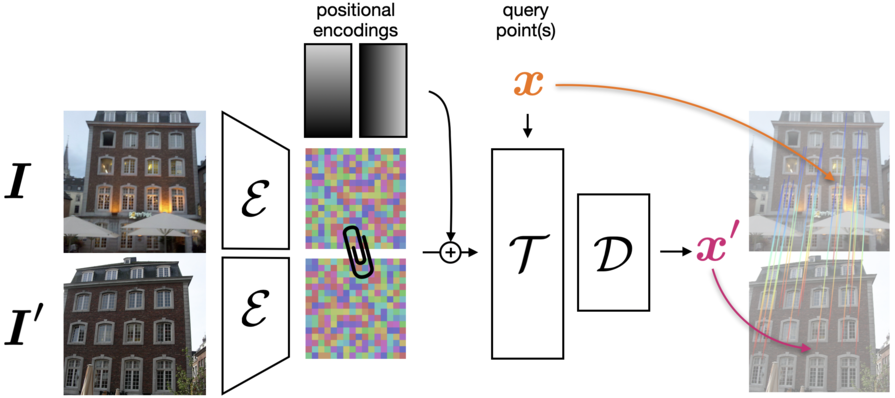}
\caption{
\textbf{The $\COTR$ architecture --}
We first process each image with a (shared) backbone CNN $\encoder$ to produce feature maps size 16x16, which we then concatenate together, and add positional encodings to form our context feature map.
The results are fed into a transformer $\mathcal{T}$, along with the query point(s) $\bx$.
The output of the transformer is decoded by a multi-layer perceptron~$\mathcal{D}$ into correspondence(s)~$\bx'$.
} %
\vspace{-1em}
\label{fig:arch}
\end{figure}

We implement $\net_{\bPhi}$ with a transformer.
Our architecture, inspired by \cite{Carion20,dosovitskiy2020image}, is illustrated in \Figure{arch}. 
We first crop and resize the input into a $256\times256$ image, and convert it into a downsampled feature map size $16 \times 16 \times 256$ with a shared CNN backbone, $\encoder$.
We then concatenate the representations for two corresponding images \emph{side by side}, forming a feature map size $16 \times 32 \times 256$, to which we add positional encoding $\pencode$ (with $N{=}256$ channels) of the coordinate function~$\positional$ (i.e. $\text{MeshGrid}(0{:}1,0{:}2)$ of size $16 {\times} 32 {\times} 2$) to produce a \emph{context feature map} $\mathbf{c}$ (of size $16 \times 32 \times 256$):
\begin{align}
\mathbf{c} = \left[\encoder(\bI)\stitchop \encoder(\bI')\right] + \pencode(\positional),
\end{align}
where $\left[ \cdot \right]$ denotes concatenation along the spatial dimension~--~a subtly important detail novel to our architecture that we discuss in greater depth later on.
We then feed the context feature map $\mathbf{c}$ to a transformer encoder $\tencoder$, and interpret its results with a transformer decoder $\tdecoder$, along with the query point $\bx$, encoded by $\pencode$ -- the positional encoder used to generate $\positional$.
We finally process the output of the transformer decoder with a fully connected layer $\decoder$ to obtain our estimate for the corresponding point, $\bx'$.
\begin{align}
\bx' = \net_{\bPhi}\left(\bx| \bI, \bI'\right) =
\decoder\left(
\tdecoder\left(
\pencode\left(\bx\right), \tencoder\left(\mathbf{c}\right)
\right)
\right).
\end{align}
For architectural details of each component please refer to \SupplementaryMaterial.

\paragraph{Importance of context concatenation}
Concatenation of the feature maps along the spatial dimension is critical, as it allows the transformer encoder $\tencoder$ to relate between locations \textit{within} the image (self-attention), and \textit{across images} (cross-attention).
Note that, to allow the encoder to distinguish between pixels in the two images, we employ a \textit{single} positional encoding for the entire concatenated feature map; see~\Fig{arch}.
We concatenate along the spatial dimension rather than the channel dimension, as the latter would create artificial relationships between features coming from the same pixel locations in each image.
Concatenation allows the features in each map to be treated in a way that is similar to words in a sentence~\cite{Vaswani17}.
The encoder then associates and relates them to discover which ones to attend to given their context -- which is arguably a more natural way to find correspondences.

\paragraph{Linear positional encoding}
\label{sec:positional}
We found it critical to use a \textit{linear} increase in frequency for the positional encoding, as opposed to the commonly used log-linear strategy~\cite{Vaswani17,Carion20}, which made our optimization unstable; see~\SupplementaryMaterial.
Hence, for a given location $\bx = [x, y]$ we write
\begin{align}
\pencode(\bx) &= \left[p_1(\bx), p_2(\bx), \dots, p_{\frac{N}{4}}(\bx)\right] ,
\\
p_k(\bx) &= \left[\sin(k\pi \bx^\top), \cos(k\pi\bx^\top)\right],
\end{align}
where $N{}={}256$ is the number of channels of the feature map.
Note that $p_k$ generates four values, so that the output of the encoder $\pencode$ is size $N$.

\paragraph{Querying multiple points}
We have introduced our framework as a function operating on a single query point, $\bx$.
However, as shown in \Fig{arch}, extending it to multiple query points is straightforward. 
We can simply input multiple queries at once, which the transformer decoder $\tdecoder$ and the decoder $\decoder$ will translate into multiple coordinates.
Importantly, while doing so, we disallow self attention among the query points in order to ensure that they are solved \textit{independently}.

\begin{figure}
\centering
\includegraphics[width=1.0\linewidth]{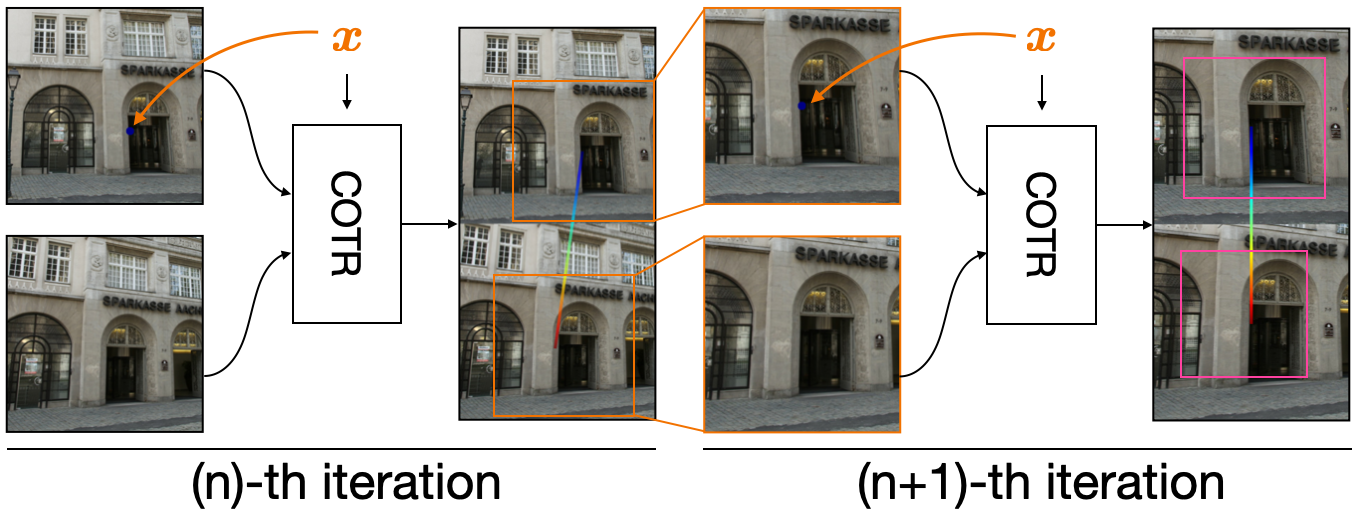}
\caption{
{\bf Recursive $\COTR$ at inference time --}
We obtain accurate correspondences by applying our functional approach recursively,
zooming into the results of the previous iteration, and running the \emph{same} network on the
pair of zoomed-in crops.
We gradually focus on the correct correspondence, with greater accuracy. %
}
\vspace{-1em}
\label{fig:infer}
\end{figure}

\subsection{Inference}
\label{sec:inference}
We next discuss how to apply our functional approach at inference time in order to obtain \textit{accurate} correspondences.

\paragraph{Inference with recursive with zoom-in}
Applying the powerful transformer attention mechanism to vision problems comes at a cost -- it requires heavily downsampled feature maps, which in our case naturally translates to poorly localized correspondences; see~\Section{ablationzoom}.
We address this by exploiting the functional nature of our approach, applying out network $\net_\Phi$ \emph{recursively}.
As shown in \Fig{infer}, we iteratively zoom into a previously estimated correspondence, on both images, in order to obtain a refined estimate. 
There is a trade-off between compute and the number of zoom-in steps.
We ablated this carefully on the validation data and settled on a zoom-in factor of two at each step, with four zoom-in steps. 
It is worth noting that multiscale refinement is common in many computer vision algorithms~\cite{liu2010sift,dosovitskiy2015flownet}, but thanks to our functional correspondence model, realizing such a multiscale inference process is not only possible, but also straightforward to implement.

\begin{figure}
\centering
\includegraphics[width=\linewidth]{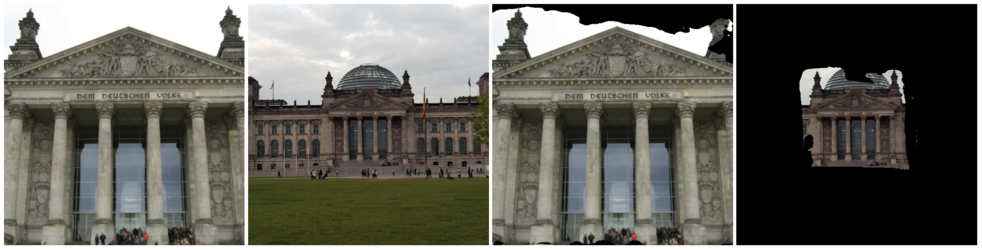}
\caption{
{\bf Estimating scale by finding co-visible regions --}~%
We show two images we wish to put in correspondence, and the estimated regions in common~--~image locations with a high cycle-consistency error are masked out.
}%
\vspace{-1em}
\label{fig:commonregion}
\end{figure}

\paragraph{Compensating for scale differences}
While matching images recursively, one must account for a potential mismatch in scale between images.
We achieve this by making the scale of the patch to crop proportional to the commonly visible regions in each image, which we compute on the first step, using the whole images.
To extract this region, we compute the cycle consistency error at the coarsest level, for every pixel, and threshold it at $\tau_\text{visible}{=}5$ pixels on the $256\times256$ image; see~\Fig{commonregion}.
In subsequent stages~--~the zoom-ins~--~we simply adjust the crop sizes over $\bI$ and $\bI'$ so that their relationship is proportional to the sum of valid pixels (the unmasked pixels in \Fig{commonregion}).

\paragraph{Dealing with images of arbitrary size}
Our network expects images of fixed $256\times256$ shape.
To process images of arbitrary size, in the initial step we simply resize (i.e.~stretch) them to $256\times256$, and estimate the initial correspondences.
In subsequent zoom-ins, we crop square patches from the original image around the estimated points, of a size commensurate with the current zoom level, and resize them to $256\times256$.
While this may seem a limitation on images with non-standard aspect ratios, our approach performs well on KITTI, which are extremely wide (3.3:1).
Moreover, we present a strategy to tile detections in Section~\ref{sec:imc2020}.

\paragraph{Discarding erroneous correspondences}
What should we do when we query a point is occluded or outside the viewport in the other image?
Similarly to our strategy to compensate for scale, we resolve this problem by simply rejecting correspondences that induce a cycle consistency error~\eq{cycle} greater than $\tau_\text{cycle}{=}5$ pixels.
Another heuristic we apply is to terminate correspondences that do not converge while zooming in.
We compute the standard deviation of the zoom-in estimates, and reject correspondences that oscillate by more than $\tau_\text{std}{=}0.02$ of the long-edge of the image.

\paragraph{Interpolating for dense correspondence}
While we could query every single point in order to obtain dense estimates, it is also possible to densify matches by computing sparse matches first, and then interpolating using barycentric weights on a Delaunay triangulation of the queries.
This interpolation can be done efficiently using a GPU rasterizer.

\subsection{Implementation details}
\label{sec:implementation}

\paragraph{Datasets}
We train our method on the MegaDepth dataset~\cite{Li18}, which provides both images and corresponding dense depth maps, generated by SfM~\cite{schoenberger2016sfm}.
These images come from photo-tourism and show large variations in appearance and viewpoint, which is required to learn invariant models.
The accuracy of the depth maps is sufficient to learn accurate local features, as demonstrated by~\cite{dusmanu2019d2,Sarlin20,Tyszkiewicz20}.
To find co-visible pairs of images we can train with, we first filter out those with no common 3D points in the SfM model. We then compute the common area between the remaining pairs of images, by projecting pixels from one image to the other.
Finally, we compute the intersection over union of the projected pixels, which accounts for different image sizes.
We keep, for each image, the 20 image pairs with the largest overlap.
This simple procedure results in a good combination of images with a mixture of high/low overlap.
We use 115 scenes for training and 1 scene for validation.

\paragraph{Implementation}
We implement our method in PyTorch~\cite{Paszke19}.
For the 
backbone $\encoder$ we use a ResNet50~\cite{He16}, initialized with weights pre-trained on ImageNet~\cite{Russakovsky15}.
We use the feature map after its fourth downsampling step (after the third residual block), which is of size $16\times16\times1024$, which we convert into $16\times16\times256$ with $1\times1$ convolutions.
For the transformer, we use 6 layers for both encoder and decoder. 
Each encoder layer contains a self-attention layer with 8 heads, and each decoder layer contains an encoder-decoder attention layer with 8 heads, but with no self-attention layers, in order to prevent query points from communicating between each other.
Finally, for the network that converts the Transformer output into coordinates, $\decoder$, we use a 3-layer MLP, with 256 units each, followed by ReLU activations.

\paragraph{On-the-fly training data generation}
We select training pairs randomly, pick a random query point in the first image, and find its corresponding point on the second image using the ground truth depth maps.
We then select a random zoom level among one of ten levels, uniformly spaced, in log scale, between 1$\times$ and 10$\times$.
We then crop a square patch at the desired zoom level, centered at the query point, from the first image, and a square patch that \emph{contains} the corresponding point in the second image.
Given this pair of crops, we sample 100 random valid correspondences across the two crops~--~if we cannot gather at least 100 valid points, we discard the pair and move to the next.

\paragraph{Staged training}
Our model is trained in three stages.
First, we freeze the pre-trained backbone $\encoder$, and train the rest of the network, for 300k iterations, with the ADAM optimizer~\cite{Kingma14}, a learning rate of $10^{-4}$, and a batch size of 24.
We then unfreeze the backbone and fine-tune everything end-to-end with a learning rate of $10^{-5}$ and a batch size of 16, to accommodate the increased memory requirements, for 2M iterations, at which point the validation loss plateaus.
Note that in the first two stages we use the whole images, resized to $256\times256$, as input, which allows us to load the entire dataset into memory.
In the third stage we introduce zoom-ins, generated as explained above, and train everything end-to-end for a further 300k iterations.
\section{Results}
We evaluate our method with four different datasets, each aimed for a different type of correspondence task. We do not perform \emph{any kind} of re-training or fine-tuning. They are:
\vspace{-.5em}
\begin{itemize}[leftmargin=*]
\setlength\itemsep{-.3em}
    \item HPatches~\cite{Balntas17}: A dataset with planar surfaces viewed under different angles/illumination settings, and ground-truth homographies.
    We use this dataset to compare against dense methods that operate on the entire image.
    \item KITTI~\cite{Geiger2013IJRR}: A dataset for autonomous driving, where the ground-truth 3D information is collected via LIDAR. 
    With this dataset we compare against dense methods on complex scenes with camera and multi-object motion.
    \item ETH3D~\cite{schoeps2017cvpr}: A dataset containing indoor and outdoor scenes captured using a hand-held camera, registered with SfM.
    As it contains video sequences, we use it to evaluate how methods perform as the baseline widens by increasing the interval between samples, following~\cite{Truong20a}.
    \item Image Matching Challenge (IMC2020)~\cite{Jin20}: A dataset and challenge containing wide-baseline stereo pairs from photo-tourism images, similar to those we use for training~(on MegaDepth).
    It takes matches as input and measures the quality the poses estimated using said matches.
    We evaluate our method on the test set
    and compare against the state of the art in sparse methods.
\end{itemize}

\begin{table}
\setlength{\tabcolsep}{6pt}
\begin{center}
\resizebox{\linewidth}{!}{
\begin{tabular}{lcccc}
\toprule
\multicolumn{1}{c}{Method}     & AEPE $\downarrow$  & PCK-1px $\uparrow$ & PCK-3px $\uparrow$ & PCK-5px $\uparrow$ \\
\midrule
\LiteFlowNet & 118.85 & 13.91 & -- & 31.64 \\
\PWCNet & 96.14  & 13.14 & -- & 37.14 \\
\DGCNet & 33.26  & 12.00 & -- & 58.06 \\
\GLUNet & 25.05  & 39.55 & 71.52 & 78.54 \\
\GOCor & 20.16  & \bf{41.55} & -- & 81.43 \\
\midrule
\COTR                &\bf{7.75}   & \underline{40.91} & \textbf{82.37} & \bf{91.10} \\
\COTRT  & \underline{7.98}   & 33.08 & \underline{77.09} & \underline{86.33} \\
\bottomrule
\end{tabular}
} %
\end{center}
\vspace{-.5em}
\caption{
{\bf Quantitative results on HPatches -- }
We report Average End Point Error (AEPE) and Percent of Correct Keypoints (PCK) with different thresholds. 
For PCK-1px and PCK-5px, we use the numbers reported in literature.
We \textbf{bold} the best method and \underline{underline} the second best.
}
\vspace{-1em}
\label{tbl:hpatches}
\end{table}
\subsection{HPatches}
We follow the evaluation protocol of~\cite{Truong20a,Truong20b},
which computes the Average End Point Error~(AEPE) for all valid pixels, and the Percentage of Correct Keypoints~(PCK) at a given reprojection error threshold~--~we use 1, 3, and 5 pixels.
Image pairs are generated taking the first (out of six) images for each scene as reference, which is matched against the other five.
We provide two results for our method:~`\COTRnobf', which uses 1,000 random query points for each image pair, and~`\COTRnobf+~Interp.', which interpolates correspondences for the remaining pixels using the strategy presented in \Section{inference}.
We report our results in~\Table{hpatches}.

Our method provides the best results, with and without interpolation, with the exception of PCK-1px, where it remains close to the best baseline.
We note that the results for this threshold should be taken with a grain of salt, as several scenes do not satisfy the planar assumption for all pixels.
To provide some evidence for this, we reproduce the results for GLU-Net~\cite{Truong20a} using the code provided by the authors to measure PCK at 3 pixels, which was not computed in the paper.~\footnote{While GLU-Net+GOCor slightly edges out GLU-Net, code was not available at the time of submission. 
}
\COTRnobf outperforms it by a significant margin.

\def \figkittiwidth {0.3}

\newcommand{\imgI}{\includegraphics[width=\figkittiwidth\linewidth]{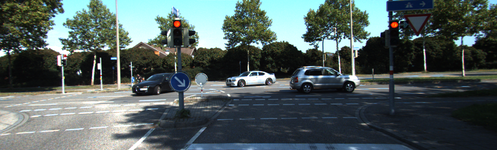}}
\newcommand{\imgII}{\includegraphics[width=\figkittiwidth\linewidth]{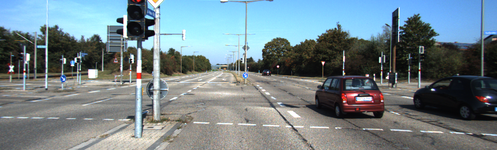}}
\newcommand{\imgIII}{\includegraphics[width=\figkittiwidth\linewidth]{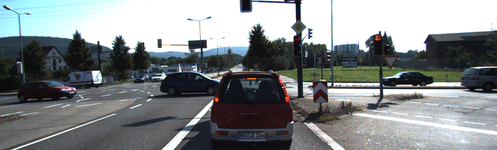}}
\newcommand{\imgIV}{\includegraphics[width=\figkittiwidth\linewidth]{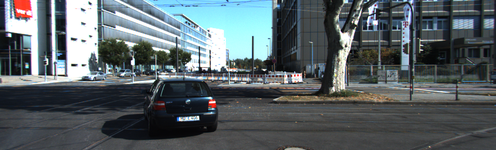}}

\newcommand{\glunetFlowI}{\includegraphics[width=\figkittiwidth\linewidth]{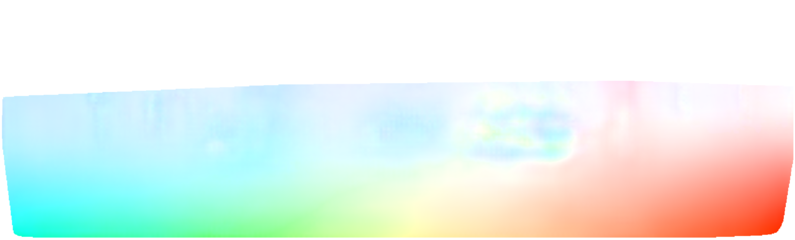}}
\newcommand{\glunetFlowII}{\includegraphics[width=\figkittiwidth\linewidth]{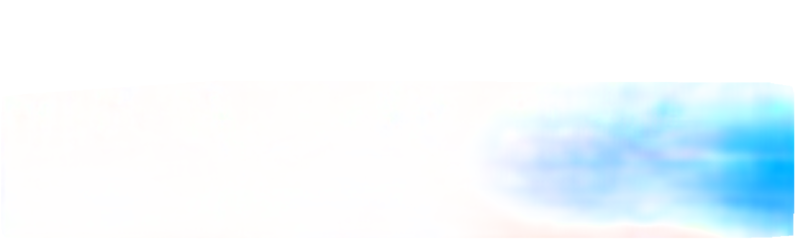}}
\newcommand{\glunetFlowIII}{\includegraphics[width=\figkittiwidth\linewidth]{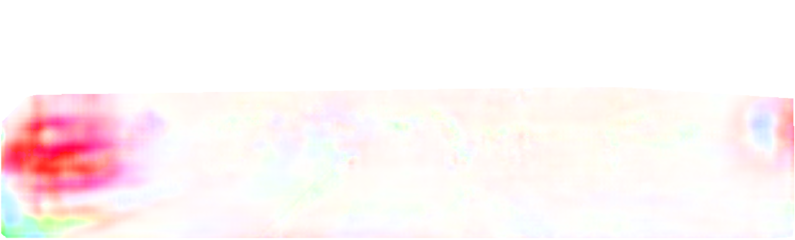}}
\newcommand{\glunetFlowIV}{\includegraphics[width=\figkittiwidth\linewidth]{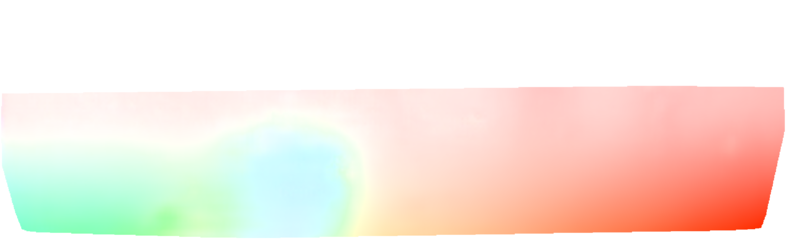}}

\newcommand{\glunetErrorI}{\includegraphics[width=\figkittiwidth\linewidth]{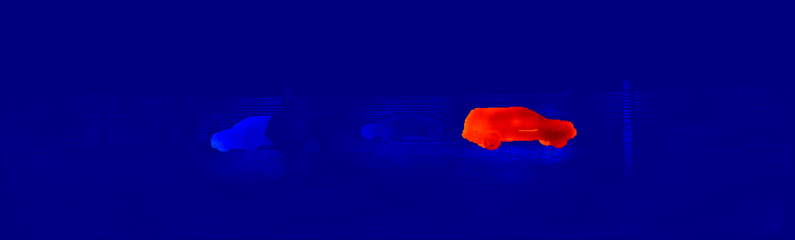}}
\newcommand{\glunetErrorII}{\includegraphics[width=\figkittiwidth\linewidth]{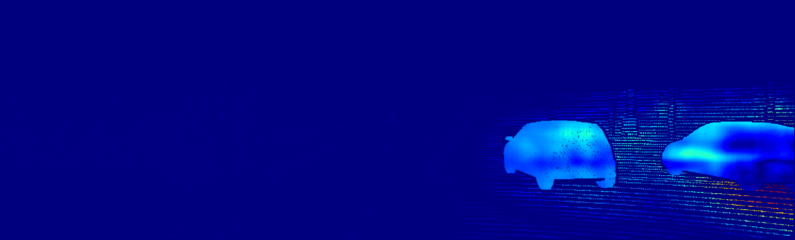}}
\newcommand{\glunetErrorIII}{\includegraphics[width=\figkittiwidth\linewidth]{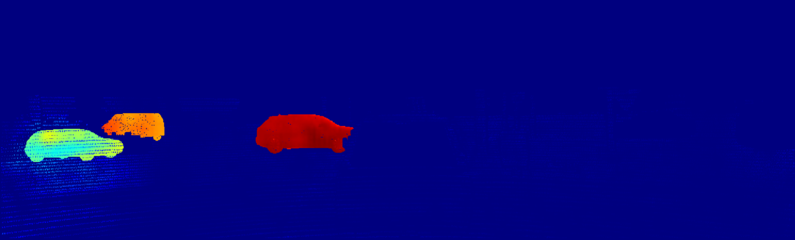}}
\newcommand{\glunetErrorIV}{\includegraphics[width=\figkittiwidth\linewidth]{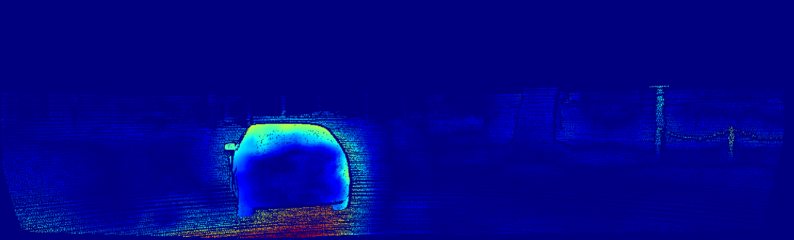}}

\newcommand{\cotrFlowI}{\includegraphics[width=\figkittiwidth\linewidth]{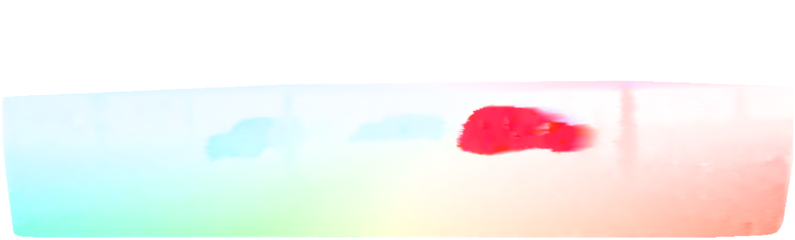}}
\newcommand{\cotrFlowII}{\includegraphics[width=\figkittiwidth\linewidth]{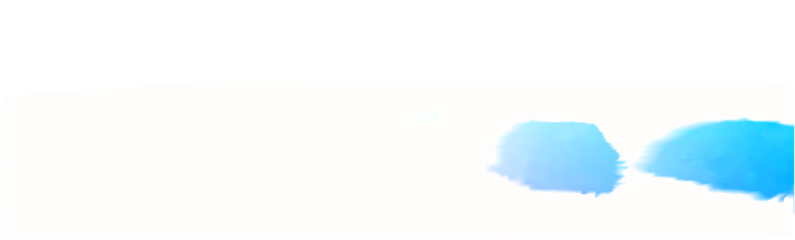}}
\newcommand{\cotrFlowIII}{\includegraphics[width=\figkittiwidth\linewidth]{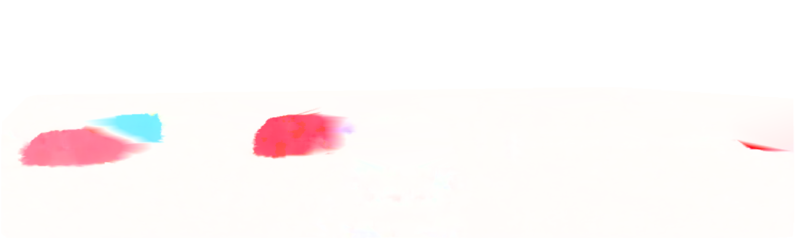}}
\newcommand{\cotrFlowIV}{\includegraphics[width=\figkittiwidth\linewidth]{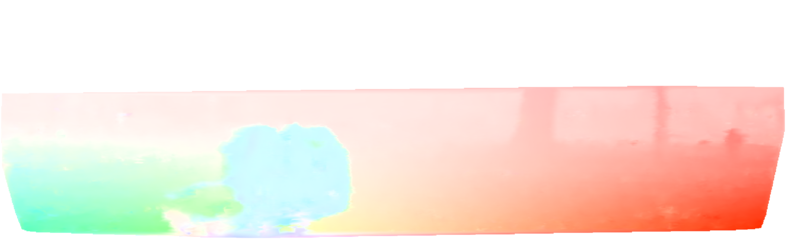}}

\newcommand{\cotrErrorI}{\includegraphics[width=\figkittiwidth\linewidth]{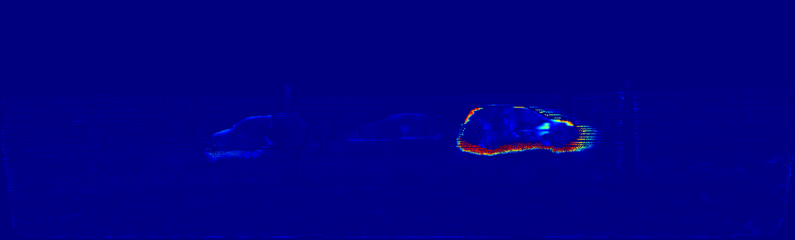}}
\newcommand{\cotrErrorII}{\includegraphics[width=\figkittiwidth\linewidth]{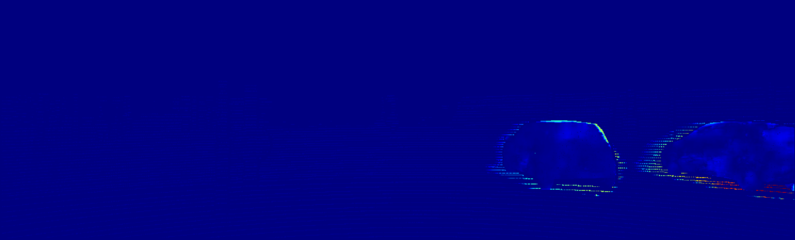}}
\newcommand{\cotrErrorIII}{\includegraphics[width=\figkittiwidth\linewidth]{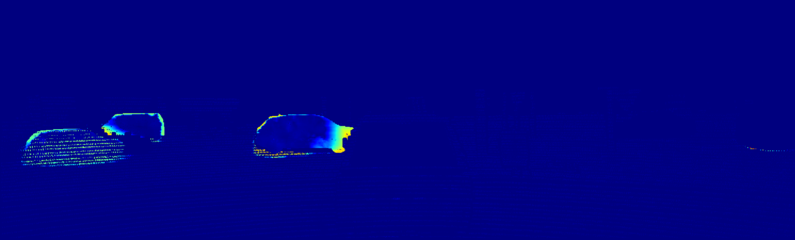}}
\newcommand{\cotrErrorIV}{\includegraphics[width=\figkittiwidth\linewidth]{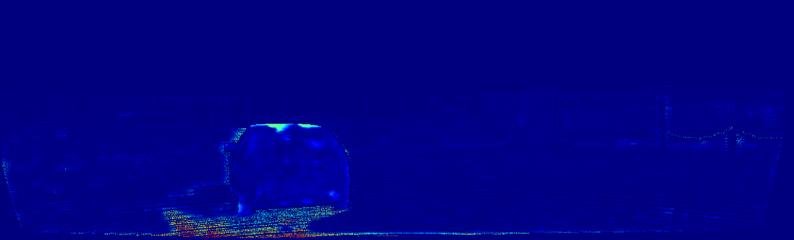}}

\begin{figure*}
\setlength{\tabcolsep}{1pt}
\begin{center}
\resizebox{\linewidth}{!}{
\begin{tabular}{c c c c c}
\imgI & \glunetFlowI & \cotrFlowI & \glunetErrorI & \cotrErrorI \\
\imgIV & \glunetFlowIV & \cotrFlowIV & \glunetErrorIV & \cotrErrorIV \\
\imgIII & \glunetFlowIII & \cotrFlowIII & \glunetErrorIII & \cotrErrorIII \\
\multirow{2}{*}{Input (shown: one image)}       & \GLUNet & \COTR (ours) & \GLUNet & \COTR (ours) \\
       & Optical flow & Optical flow & Error map & Error map \\

\end{tabular}
} %
\end{center}
\vspace{-.5em}
\caption{
{\bf Qualitative examples on KITTI --}
We show the optical flow and its corresponding error map (``jet'' color scheme) for three examples from KITTI-2015, with GLU-Net~\cite{Truong20a} as a baseline. 
\COTRnobf successfully recovers both the global motion in the scene, and the movement of individual objects, even when nearby cars move in opposite directions (top) or partially occlude each other (bottom).
}
\vspace{-1em}
\label{fig:kitti}
\end{figure*}
\subsection{KITTI}

\begin{table}
\setlength{\tabcolsep}{6pt}
\begin{center}
\resizebox{\linewidth}{!}{
\begin{tabular}{lcccccc}
\toprule
\multicolumn{1}{c}{\multirow{2}{*}[-2pt]{Method}} & \multicolumn{2}{c}{KITTI-2012}                            & \multicolumn{2}{c}{KITTI-2015}                            \\
\cmidrule(r){2-3}
\cmidrule(r){4-5}
& \multicolumn{1}{c}{AEPE$\downarrow$} & \multicolumn{1}{c}{Fl.{[}\%{]$\downarrow$}} & \multicolumn{1}{c}{AEPE$\downarrow$} & \multicolumn{1}{c}{Fl.{[}\%{]$\downarrow$}} \\
\midrule
\LiteFlowNet & 4.00  & 17.47  & 10.39  & 28.50 \\
\PWCNet      & 4.14  & 20.28  & 10.35  & 33.67 \\
\DGCNet      & 8.50  & 32.28  & 14.97  & 50.98 \\
\GLUNet      & 3.34  & 18.93  & 9.79   & 37.52 \\
\RAFT        & \underline{2.15}    & \underline{9.30}    & \underline{5.04}   & 17.8 \\
\GOCor       & 2.68  & 15.43  & 6.68   & 27.57 \\
\midrule
\COTR\footnotemark       & \textbf{1.28}  & \textbf{7.36}  & \textbf{2.62}  & \textbf{9.92}  \\
\COTRT\textcolor{red}{\textsuperscript{3}}      & 2.26  & 10.50  & 6.12   & \underline{16.90} \\
\bottomrule
\end{tabular}
} %
\end{center}
\vspace{-.5em}
\caption{
{\bf Quantitative results on KITTI --}
We report the Average End Point Error (AEPE) and the flow outlier ratio (`Fl') on the 2012 and 2015 versions of the KITTI dataset. 
Our method outperforms most baselines, with the interpolated version being on par with RAFT, and slightly edging out GLU-Net+GOCor.
}
\vspace{-1em}
\label{tbl:kitti}
\end{table}

To evaluate our method in an environment more complex than simple planar scenes, we use the KITTI dataset~\cite{Menze15, Menze18}.
Following~\cite{Truong20b,Teed20}, we use the training split for this evaluation, as ground-truth for the test split remains private~--~\textit{all} methods, including ours, were trained on a separate dataset.
We report results both in terms of AEPE, and `Fl.'~--~the percentage of optical flow outliers.
\footnotetext{We filter out points that do not satisfy the cycle-consistency constraint, thus the results are not directly comparable.}
As KITTI images are large, we randomly sample 40,000 points per image pair, from the regions covered by valid ground truth.

We report the results on both KITTI-2012 and KITTI-2015 in \Table{kitti}.
Our method outperforms all the baselines by a large margin.
Note that the interpolated version also performs similarly to the state of the art, slightly better in terms of flow accuracy, and slightly worse in terms of AEPE, compared to RAFT~\cite{Teed20}.
It is important to understand here that, while \COTRnobf provides a drastic improvement over compared methods, we are evaluating only on points where \COTRnobf returns confident results, which is about 81.8\% of the queried locations~--~among the 18.2\% of rejected queries, 67.8\% fall out of the borders of the other image, which indicates that our filtering is reasonable.
This shows that \COTRnobf provides highly accurate results in the points we query and retrieve estimates for, and is currently limited by the interpolation strategy.
This suggests that improved interpolation strategies based on CNNs, such as those used in \cite{Meshry19}, would be a promising direction for future research.

In \Fig{kitti} we further highlight cases where our method shows clear advantages over the competitors~--~we see that the objects in motion, \ie, cars, result in high errors with GLU-Net, which is biased towards a single, \textit{global} motion.
Our method, on the other hand, successfully recovers the flow fields for these cases as well, with minor errors at the boundaries, due to interpolation.
These examples clearly demonstrate the role that attention plays when estimating correspondences on scenes with moving objects.

Finally, we stress that while our method is trained on MegaDepth, an urban dataset exhibiting only global, rigid motion, for which ground truth is only available on stationary objects (mostly building facades), our method proves capable of recovering the motion of objects moving in different directions; see~\Fig{kitti}, bottom.
In other words, it learns to find \emph{precise, local correspondences} within images, rather than global motion.

\begin{table}
\setlength{\tabcolsep}{6pt}
\begin{center}
\resizebox{\linewidth}{!}{
\begin{tabular}{lccccccc}
\toprule
\multicolumn{1}{c}{\multirow{2}{*}[-2pt]{Method}}& \multicolumn{7}{c}{AEPE $\downarrow$} \\
\cmidrule(r){2-8}
 & rate=3 & \multicolumn{1}{r}{rate=5} & rate=7 & rate=9 & rate=11 & rate=13 & rate=15 \\
\midrule
\LiteFlowNet                & \textbf{1.66} & 2.58                       & 6.05          & 12.95         & 29.67         & 52.41         & 74.96         \\
\PWCNet                    & 1.75          & 2.10                       & 3.21          & 5.59          & 14.35         & 27.49         & 43.41         \\
\DGCNet                    & 2.49          & 3.28                       & 4.18          & 5.35          & 6.78          & 9.02          & 12.23         \\
\GLUNet                    & 1.98          & 2.54                       & 3.49          & 4.24          & 5.61          & 7.55          & 10.78         \\
\RAFT                    & 1.92          & 2.12                       & 2.33          & 2.58          & 3.90          & 8.63          & 13.74         \\
\midrule
\COTR     & \textbf{1.66} & \textbf{1.82}  & \textbf{1.97} & \textbf{2.13} & \textbf{2.27} & \textbf{2.41} & \textbf{2.61} \\
\COTRT    & \underline{1.71} & \underline{1.92} & \underline{2.16} & \underline{2.47} & \underline{2.85} & \underline{3.23} & \underline{3.76}            \\
\bottomrule
\end{tabular}
}
\end{center}
\vspace{-.5em}
\caption{
{\bf Quantitative results for ETH3D --}
We report the Average End Point Error (AEPE) at different sampling ``rates'' (frame intervals). 
Our method performs significantly better as the rate increases and the problem becomes more difficult.
}
\vspace{-1em}
\label{tbl:eth3d}
\end{table}
\subsection{ETH3D}
We also report results on the ETH3D dataset, following~\cite{Truong20a,Truong20b}.
This task is closer to the `sparse' scenario, as performance is only evaluated on pixels corresponding to SfM locations with valid ground truth, which are far fewer than for HPatches or KITTI.
We summarize the results in terms of AEPE in \Table{eth3d}, sampling pairs of images with an increasing number of frames between them (the sampling ``rate''), which correlates with baseline and, thus, difficulty.
Our method produces the most accurate correspondences for every setting, tied with LiteFlowNet~\cite{Hui18} at a 3-frame difference, and drastically outperforms every method as the baseline increases\footnote{We could not report exact numbers for GLU-Net+GOCor as they were not reported, and their implementation was not yet publicly available at the time of submission, but our method should comfortably outperform it in every setting; see~\cite{Truong20b}, Fig~4.}; see qualitative results in \Fig{eth3d}.

\newcommand{\cotrethI}{\includegraphics[height=0.25\linewidth]{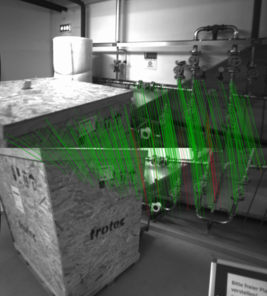}}
\newcommand{\cotrethII}{\includegraphics[height=0.25\linewidth]{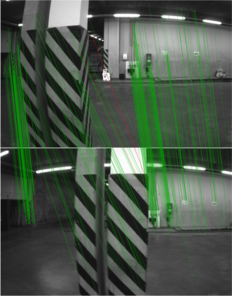}}
\newcommand{\cotrethIII}{\includegraphics[height=0.25\linewidth]{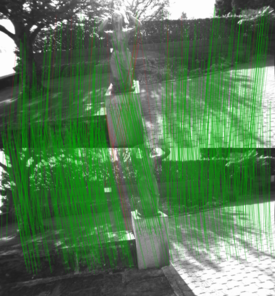}}
\newcommand{\cotrethIV}{\includegraphics[height=0.25\linewidth]{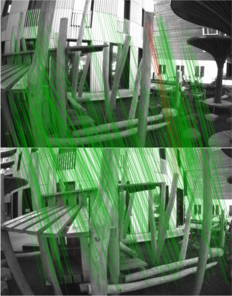}}

\newcommand{\glunetethI}{\includegraphics[height=0.25\linewidth]{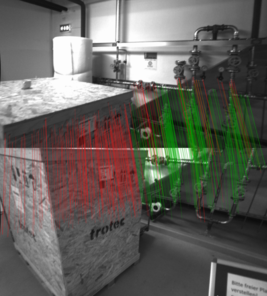}}
\newcommand{\glunetethII}{\includegraphics[height=0.25\linewidth]{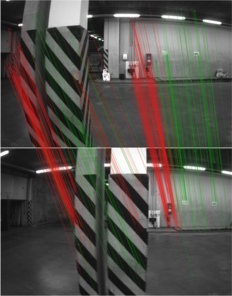}}
\newcommand{\glunetethIII}{\includegraphics[height=0.25\linewidth]{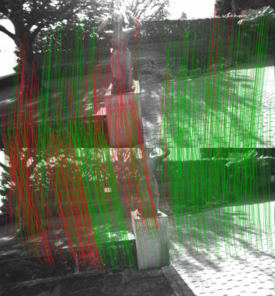}}
\newcommand{\glunetethIV}{\includegraphics[height=0.25\linewidth]{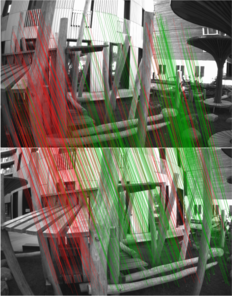}}

\begin{figure}[t]
\tiny
\setlength{\tabcolsep}{0pt}
\begin{center}
\resizebox{\linewidth}{!}{
\begin{tabular}{cccc}
 \glunetethI & \cotrethI & \glunetethIV & \cotrethIV \\
 GLU-Net~\cite{Truong20a} & \COTR & GLU-Net~\cite{Truong20a} & \COTR \\
 \cmidrule(lr){1-2} 
 \cmidrule(lr){3-4} 
 \multicolumn{2}{c}{Indoors} & \multicolumn{2}{c}{Outdoors} \\
\end{tabular}
} %
\end{center}
\vspace{-1em}
\caption{
{\bf Qualitative examples on ETH3D --}
We show results for GLU-Net~\cite{Truong20a} and \COTRnobf for two examples, one indoors and one outdoors.
Correspondences are drawn in {\bf \textcolor{eth3d_green}{green}} if their reprojection error is below 10 pixels, and {\bf \textcolor{red}{red}} otherwise.
}
\vspace{-1em}
\label{fig:eth3d}
\end{figure}

\subsection{Image Matching Challenge}
\label{sec:imc2020}
Accurate, 6-DOF pose estimation in unconstrained urban scenarios remains too challenging a problem for dense methods. 
We evaluate our method on a popular challenge for pose estimation with local features, which measures performance in terms of the quality of the estimated poses, in terms of mean average accuracy (mAA) at a 5$^\circ$ and 10$^\circ$ error threshold; see~\cite{Jin20} for details.

We focus on the stereo task.\footnote{The challenge features two tracks: stereo, and multi-view (SfM).
Our approach works on arbitrary locations and has no notion of `keypoints' (we use \emph{random} points).
For this reason, we do not consider the multiview task, as SfM requires~``stable'' points to generate 3D landmarks.
We plan to re-train the model and explore its use on keypoint locations in the future.}
As this dataset contains images with unconstrained aspect ratios, instead of stretching the image before the first zoom level, we simply resize the short-edge to $256$ and tile our coarse, image-level estimates~--~\eg an image with 2:1 aspect ratio would invoke two tiling instances.
If this process generates overlapping tiles (\eg with a 4:3 aspect ratio), we choose the estimate that gives best cycle consistency among them.
We pair our method with DEGENSAC~\cite{Chum05} to retrieve the final pose, as recommended by~\cite{Jin20} and done by most participants.

\newcommand{\cotrimwI}{\includegraphics[height=0.5\linewidth]{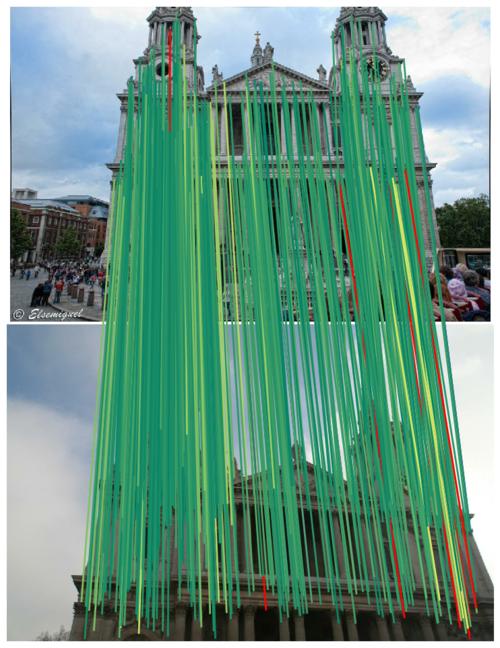}}
\newcommand{\cotrimwII}{\includegraphics[height=0.5\linewidth]{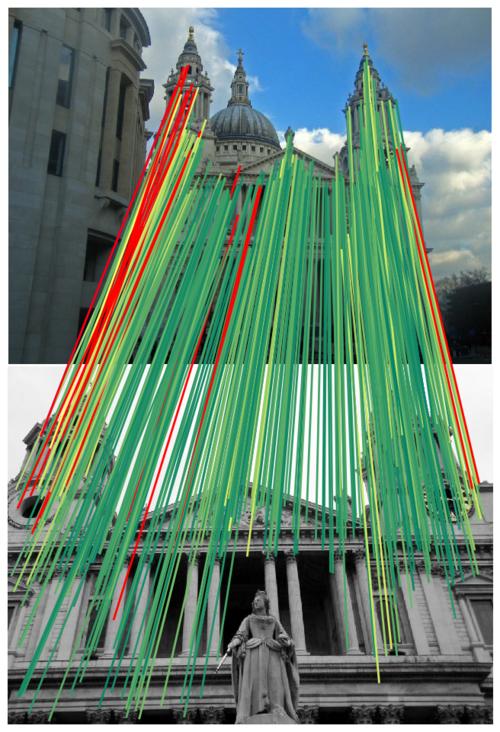}}
\newcommand{\cotrimwIII}{\includegraphics[height=0.5\linewidth]{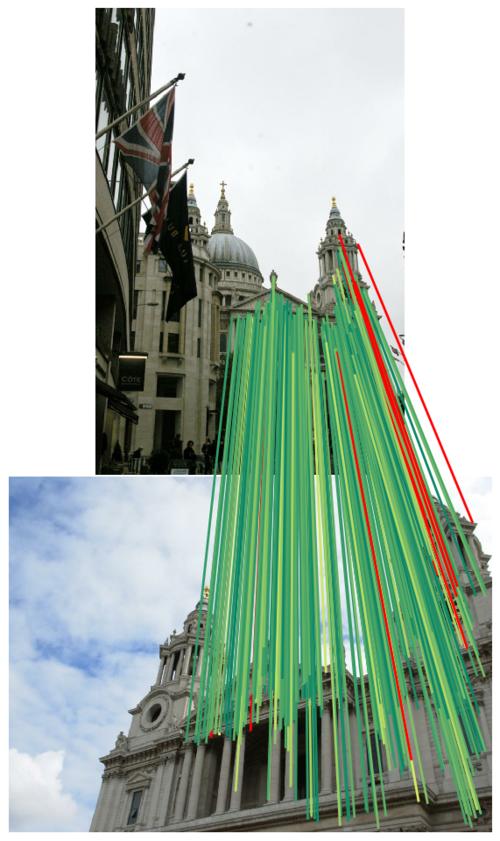}}

\begin{figure}[t]
\setlength{\tabcolsep}{0pt}
\def\arraystretch{0}
\begin{center}
\resizebox{\linewidth}{!}{
\begin{tabular}{ccc}
\cotrimwI & \cotrimwII & \cotrimwIII
\end{tabular}
} %
\end{center}
\vspace{-.5em}
\caption{
{\bf Qualitative examples for IMC2020 --}
We visualize the matches produced by \COTRnobf (with $N=512$) for some stereo pairs in the Image Matching Challenge dataset. Matches are coloured {\bf \textcolor{red}{red}} to {\bf \textcolor{eth3d_green}{green}}, according to their reprojection error (high to low).
}
\label{fig:imw}
\end{figure}
\begin{table}
\setlength{\tabcolsep}{2pt}
\begin{center}
\resizebox{\linewidth}{!}{
\begin{tabular}{l lccc}
\toprule
&\multicolumn{1}{c}{Method} & Num. Inl.$\uparrow$ & mAA(5$^\circ$)$\uparrow$ &  mAA(10$^\circ$)$\uparrow$ \\
\midrule
\multirow{4}{*}{\rotatebox[origin=l]{90}{\small 8k-keypoints}}
&\leaderboard{https://www.cs.ubc.ca/research/image-matching-challenge/2020/submissions/sid-00591-guided-sift8k_8k_hardnet-epoch2} DoG~\cite{Lowe04}+HardNet~\cite{mishchuk2017working}+ModifiedGuidedMatching & 762.0 & \textbf{0.476} & \textbf{0.611} \\
&\leaderboard{https://www.cs.ubc.ca/research/image-matching-challenge/2020/submissions/sid-00624-sift8k_8k_hardnet-epoch2-trained-from-scratch+gm/} DoG~\cite{Lowe04}+HardNet~\cite{mishchuk2017working}+OANet~\cite{zhang2019oanet}+GuidedMatching & 765.3 & \underline{0.471} & \underline{0.603} \\
&\leaderboard{https://www.cs.ubc.ca/research/image-matching-challenge/2020/submissions/sid-00610-hardnet-upright-8k-adalam/} DoG~\cite{Lowe04}+HardNet~\cite{mishchuk2017working}+AdaLAM~\cite{cavalli2020adalam}+DEGENSAC~\cite{Chum05} & 627.7 & 0.460 & 0.583 \\
&\leaderboard{https://www.cs.ubc.ca/research/image-matching-challenge/2020/submissions/sid-00621-results/} DoG~\cite{Lowe04}+HardNet8~\cite{pultar2020improving}+PCA+BatchSampling+DEGENSAC~\cite{Chum05} & 583.1 & 0.464 & 0.590 \\

\midrule
\multirow{7}{*}{\rotatebox[origin=l]{90}{\small 2k-keypoints}}
&\leaderboard{https://www.cs.ubc.ca/research/image-matching-challenge/2020/submissions/sid-00612-sp-k2048-nms4-refine2-r1600forcecubic-down128-masked-d.001-adapt50_sg-t.2-it150_degensac-th1.1} SP~\cite{Detone18}+SG~\cite{Sarlin20}+DEGENSAC~\cite{Chum05}+SemSeg+HAdapt & (441.5) & (0.452) & (0.590) \\
&\leaderboard{https://www.cs.ubc.ca/research/image-matching-challenge/2020/submissions/sid-00603-sp-k2048-nms3-refine2-r1600forcecubic-down128-masked-d.001_sg-t.2-it150_degensac-th1.2} SP~\cite{Detone18}+SG~\cite{Sarlin20}+DEGENSAC~\cite{Chum05}+SemSeg & (404.7) & (0.429) & (0.568) \\
&\leaderboard{https://www.cs.ubc.ca/research/image-matching-challenge/2020/submissions/sid-00589-sp-k2048-nms3-refine2-r1600forcecubic-down128_sg-t.2-it150_degensac-th1.2} SP~\cite{Detone18}+SG~\cite{Sarlin20}+DEGENSAC~\cite{Chum05} & 320.5 & 0.416 & 0.552 \\
&\leaderboard{https://www.cs.ubc.ca/research/image-matching-challenge/2020/submissions/sid-00708-disk-cc-continued-20-imsize-1024-nms-3-nump-2048-stereo-degensac-th-0.75-rt-0.95/} DISK~\cite{Tyszkiewicz20}+DEGENSAC~\cite{Chum05} & 404.2 & 0.388 & 0.513 \\
&\leaderboard{https://www.cs.ubc.ca/research/image-matching-challenge/2020/submissions/sid-00608-sift2k_2048_hardnet64-train-all-l2-138000-matched/} DoG~\cite{Lowe04}+HardNet~\cite{mishchuk2017working}+CustomMatch+DGNSC~\cite{Chum05} & 245.4 & 0.369 & 0.492 \\
&\leaderboard{https://www.cs.ubc.ca/research/image-matching-challenge/2020/submissions/sid-00035-hardnet-upright-magsac-2k/} DoG~\cite{Lowe04}+HardNet~\cite{mishchuk2017working}+MAGSAC~\cite{barath2019magsac} & 181.8 & 0.318 & 0.438 \\
&\leaderboard{https://www.cs.ubc.ca/research/image-matching-challenge/2020/submissions/sid-00706-logpolar-upright-2k-both-degensac-no-flann/} DoG~\cite{Lowe04}+LogPolarDesc~\cite{ebel2019beyond}+DEGENSAC~\cite{Chum05} & 162.2 & 0.333 & 0.457 \\[0.5ex]
\midrule
\multirow{6}{*}{\rotatebox[origin=l]{90}{\small Ours}}
&\COTR+DEGENSAC~\cite{Chum05} ($N=2048$) & 1676.6 & 0.444 & 0.580 \\
&\COTR+DEGENSAC~\cite{Chum05} ($N=1024$) & 840.3 & 0.435 & 0.571 \\
&\COTR+DEGENSAC~\cite{Chum05} ($N=512$) & 421.3 & 0.418 & 0.555 \\
&\COTR+DEGENSAC~\cite{Chum05} ($N=256$) & 211.7 & 0.392 & 0.529 \\
&\COTR+DEGENSAC~\cite{Chum05} ($N=128$) & 106.8 & 0.356 & 0.492 \\
\bottomrule
\end{tabular}
} %
\end{center}
\vspace{-.5em}
\caption{
{\bf Stereo performance on IMC2020 --}
We report mean Average Accuracy (mAA) at 5$^\circ$ and 10$^\circ$, and the number of inlier matches, for the top IMC2020 entries, on all test scenes.
We highlight the best method in \textbf{bold} and \underline{underline} the second-best.
We exclude entries with components specifically tailored to the challenge, which are enclosed in parentheses, but report them for completeness.
Finally, we report results with different number of matches ($N$) under pure \COTRnobf and one entry with 2048 keypoints under \COTRnobf guided matching. Pure \COTRnobf outperforms \emph{all methods in the 2k-keypoints category} (other than those specifically excluded) with as few as 512 matches per image.
With \textcolor{magenta}{\small \URLicon} we indicate clickable URLs to the leaderboard webpage.
}
\label{tbl:imw}
\end{table}

We summarize the results in \Table{imw}. 
We consider the top performers in the 2020 challenge (a total of 228 entries can be found in the leaderboards [\href{https://www.cs.ubc.ca/research/image-matching-challenge/leaderboard/}{link}]).
As the challenge places a limit on the number of keypoints, instead of matches, we consider both categories (up to 2k and up to 8k keypoints per image), for fairness~--~note that our method has no notion of \emph{keypoints}, instead, we query at \emph{random} locations.\footnote{While we limit the number of matches for each image pair, because we use random points for each pair, the number of points we use \emph{per image} may grow very large. Hence, our method does not fit into the `traditional' image matching pipeline, requiring additional considerations to use this benchmark; we thank the organizers for accommodating our request.}

With 2k matches and excluding the methods that feature semantic masking~--~a heuristic employed in the challenge by some participants to filter out keypoints on transient structures such as the sky or pedestrians~--~\COTRnobf \textit{ranks second overall}.
These results showcase the robustness and generality of our method, considering that it was not trained specifically to solve wide-baseline stereo problems.
In contrast, the other top entries are engineered towards this specific application.
We also provide results lowering the cap on the number of matches (see $N$ in \Table{imw}), showing that our method outperforms vanilla SuperGlue~\cite{Sarlin20} (the winner of the 2k-keypoint category) with as few as 512 input matches, and DISK~\cite{Tyszkiewicz20} (the runner-up) with as few as 256 input matches.
Qualitative examples on IMC are illustrated in~\Fig{imw}.

\newcommand{\cokeI}{\includegraphics[height=0.5\linewidth]{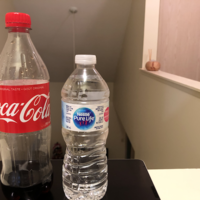}}
\newcommand{\cokeII}{\includegraphics[height=0.5\linewidth]{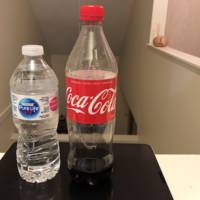}}
\newcommand{\cokeIII}{\includegraphics[height=0.5\linewidth]{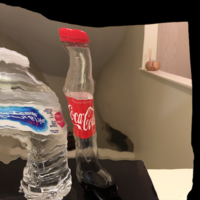}}
\newcommand{\cokeIV}{\includegraphics[height=0.5\linewidth]{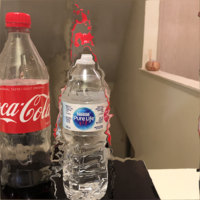}}

\begin{figure}
\setlength{\tabcolsep}{0pt}
\begin{center}
\resizebox{\linewidth}{!}{
\begin{tabular}{cccc}
 \cokeII & \cokeI & \cokeIII & \cokeIV \\
 Source & Target & \GLUNet & \COTR \\
\end{tabular}
} %
\end{center}
\vspace{-.5em}
\caption{
{\bf Object-centric scenes -- }
We compute dense correspondences with GLU-Net~\cite{Truong20a} and \COTRnobf, and warp the source image to the target image with the resulting flows.
GLU-Net fails to capture the bottles being swapped, contrary to our method.
}
\label{fig:object}
\end{figure}
\subsection{Object-centric scenes}
While our evaluation focuses on outdoor scenes, our models can be applied to very different images, such as those picturing objects.
We show one such example in \Fig{object}, where \COTRnobf successfully estimates dense correspondences for two of objects moving in different directions~--~despite the fact that this data looks \textit{nothing} alike the images it was trained with.
This shows the generality of our approach.

\subsection{Ablation studies}
\label{sec:ablation}

\paragraph{Filtering}
We validate the effectiveness of filtering out bad correspondences (Section~\ref{sec:inference}) on the ETH3D dataset, where it improves AEPE by roughly 5\% relative.
More importantly, it effectively removes correspondences with a potentially high error.
This allows the dense interpolation step to produce better results.
We find that on average 1.2\% of the correspondences are filtered out on this dataset~--~below 1\% up to `rate=9', gradually increasing until 3.65\% at `rate=15'.

\paragraph{On the role of the transformer}
\label{sec:ablationmlp}
Transformers are powerful attention mechanisms, but also costly.
It is fair to wonder whether a simpler approach would suffice.
We explore the use of MLPs in place of transformers, forming a pipeline similar to \cite{Halimi20}, and train such a variant~--~see~\SupplementaryMaterial for details.
In \Fig{vsmlp}, we see that the MLP yields globally-smooth estimates, as expected, which fail to model the discontinuities that occur due to 3D geometry. 
On the other hand, \COTRnobf with the transformer successfully aligns source and target even when such discontinuities exist.

\def \imgmlph {0.45}

\newcommand{\imgImlpI}{\includegraphics[height=\imgmlph\linewidth]{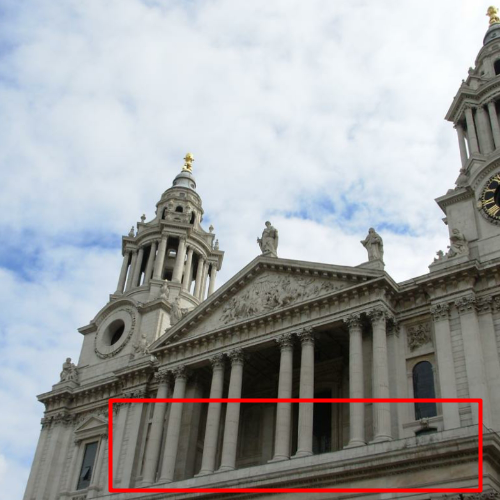}}
\newcommand{\imgIImlpI}{\includegraphics[height=\imgmlph\linewidth]{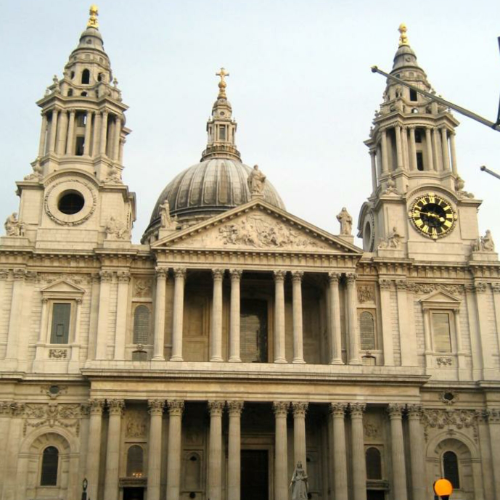}}
\newcommand{\cotrmlpI}{\includegraphics[height=\imgmlph\linewidth]{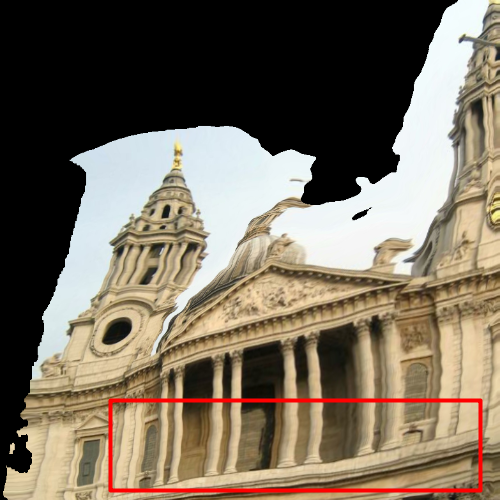}}
\newcommand{\comlpmlpI}{\includegraphics[height=\imgmlph\linewidth]{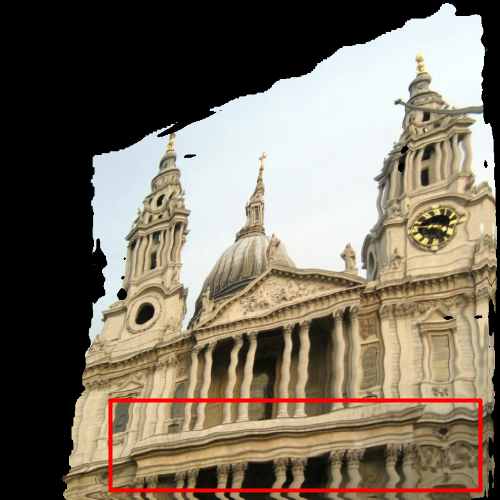}}

\newcommand{\imgImlpII}{\includegraphics[height=\imgmlph\linewidth]{fig/vsmlp/imgs/02_im1.png}}
\newcommand{\imgIImlpII}{\includegraphics[height=\imgmlph\linewidth]{fig/vsmlp/imgs/02_im2.png}}
\newcommand{\cotrmlpII}{\includegraphics[height=\imgmlph\linewidth]{fig/vsmlp/imgs/02_warp_cotr.png}}
\newcommand{\comlpmlpII}{\includegraphics[height=\imgmlph\linewidth]{fig/vsmlp/imgs/02_warp_comlp.png}}

\newcommand{\imgImlpIV}{\includegraphics[height=\imgmlph\linewidth]{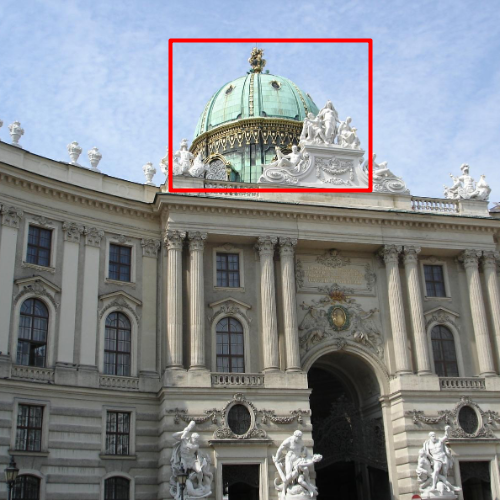}}
\newcommand{\imgIImlpIV}{\includegraphics[height=\imgmlph\linewidth]{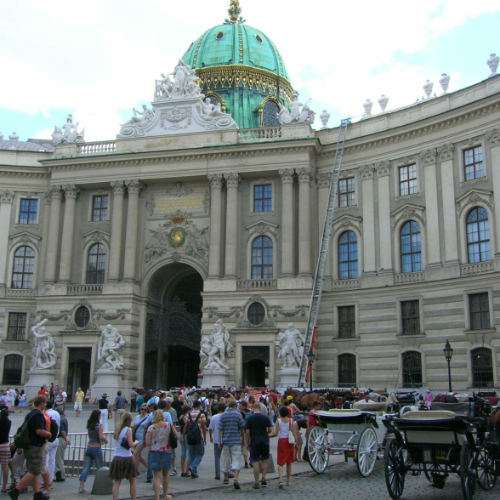}}
\newcommand{\cotrmlpIV}{\includegraphics[height=\imgmlph\linewidth]{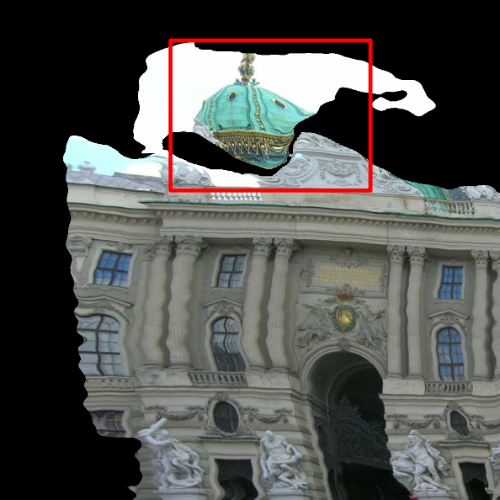}}
\newcommand{\comlpmlpIV}{\includegraphics[height=\imgmlph\linewidth]{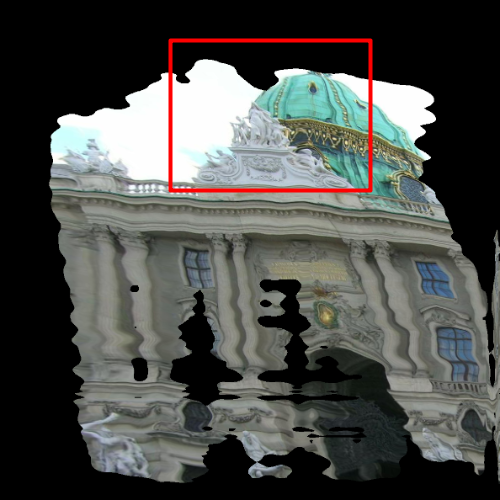}}

\begin{figure}
\setlength{\tabcolsep}{0pt}
\begin{center}
\resizebox{\linewidth}{!}{
\begin{tabular}{cccc}
 \imgIImlpI & \imgImlpI & \comlpmlpI & \cotrmlpI \\
 \imgIImlpIV & \imgImlpIV & \comlpmlpIV & \cotrmlpIV \\
 Source & Target & With MLP & With transformer \\
\end{tabular}
} %
\vspace{-.5em}
\end{center}
\caption{
{\bf Transformer vs MLPs --}
We show examples of warping the source image onto the target image using estimated dense flows, for two stereo pairs from (top row) the Image Matching Challenge test set and (bottom row) scene `0360' of the MegaDepth dataset, which was not used for training nor validation.
We use both \COTRnobf and a variant replacing the transformer with MLPs.
We compute dense correspondences at the coarsest level (for ease of illustration), and use them to warp the source image onto the target image.
Note how the MLP cannot capture the various discontinuities that occur due to the non-planar 3D structure, and instead tries to solve the problem with planar warps which produce clear artefacts (top row), and is also unable to match the dome of the palace (bottom row).
Our method with the transformer (bottom row) succeeds in both.
}
\label{fig:vsmlp}
\end{figure}
\begin{figure}
\centering
\includegraphics[width=0.85\linewidth]{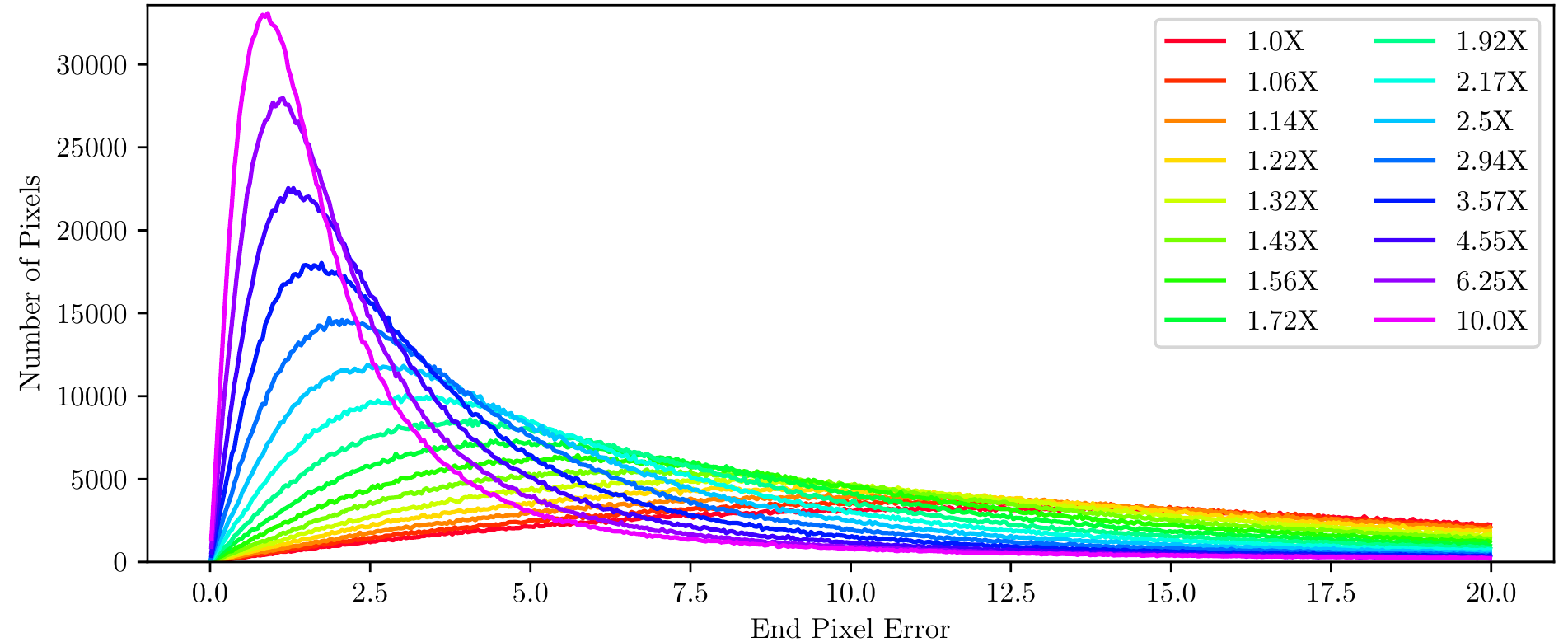}
\caption{
{\bf Zooming --}
We plot the distribution of the end pixel error (EPE) at different zoom-in levels, on the HPatches dataset.
The error clearly decreases as more zoom is applied.
}
\label{fig:zoom_error}
\end{figure}

\paragraph{Zooming}
\label{sec:ablationzoom}
To evaluate how our zooming strategy affects the localization accuracy of the correspondences, we measure the errors in the estimation at each zoom level, in pixels.
We use the HPatches dataset, with more granularity than we use for inference, and display the histogram of pixels errors at each zoom level in \Fig{zoom_error}.
As we zoom-in, the distribution shifts to the left and gets squeezed, yielding more accurate estimates.
While zooming in more is nearly always beneficial, we found empirically that four zoom-ins with a factor of two at each zoom provides a good balance between compute and accuracy.

\section{Conclusions and future work}
We introduced a functional network for image correspondence that is capable to address \textit{both} sparse and dense matching problems.
Through a novel architecture and recursive inference scheme, it achieves performance on par or above the state of the art on HPatches, KITTI, ETH3D, and one scene from IMC2020.
As future work, in addition to the improvements we have suggested throughout the paper, we intend to explore the application of \COTRnobf to semantic and multi-modal matching, and incorporate refinement techniques to further improve the quality of its dense estimates.

\section*{Acknowledgements}
This work was supported by the Natural Sciences and Engineering Research Council of Canada (NSERC) Discovery Grant, Google's Visual Positioning System, Compute Canada, and Advanced Research Computing at the University of British Columbia.

{
    \balance
    \small
    \bibliographystyle{ieee_fullname}
    \bibliography{macros,main}
}
\clearpage
\appendix
\renewcommand\thefigure{\Alph{figure}}
\setcounter{figure}{0}
\renewcommand\thetable{\Alph{table}}
\setcounter{table}{0}
\setcounter{footnote}{0}
\twocolumn[
\centering
\Large
\textbf{\COTR: Correspondence Transformer for Matching Across Images} \\
\vspace{0.5em}Supplementary Material \\
\vspace{1.0em}
]

\section{Compute}

The functional (and recursive) nature of our approach, coupled with the use of a transformer, means that our method has significant compute requirements.
Our currently non-optimized prototype implementation queries \textit{one} point at a time, and achieves 35 correspondences per second on a NVIDIA RTX~3090 GPU.
This limitation could be addressed by careful engineering in terms of tiling and batching.
Our preliminary experiments show no significant drop in performance when we query different points inside a given crop~--~we could thus potentially process any queries at the coarsest level in a single operation, and drastically reduce the number of operations in the zoom-ins (depending on how many queries overlap in a given crop).
We expect this will speed up inference drastically.
In addition to batching the queries at inference time, we plan to explore its use on non-random points (such as keypoints) and advanced interpolation techniques.

\section{Log-linear vs Linear}

Here, we empirically demonstrate that linear positional encoding is important.
We train two \COTR models with different positional encoding strategies; see~\Section{positional}.
One model uses log-linear increase in the frequency of the sine/cosine function, and the other uses linear increase instead.
\Fig{linear} shows that \COTR successfully converges using the linear increase strategy.
However, as shown in \Fig{log}, \COTR fails to converge with the commonly used log-linear strategy~\cite{Vaswani17,Carion20}.
We suspect that this is because the task of finding correspondences does not involve very high frequency components, but further investigation is necessary and is left as future work.

\newcommand{\lineartrain}{\includegraphics[height=0.5\linewidth]{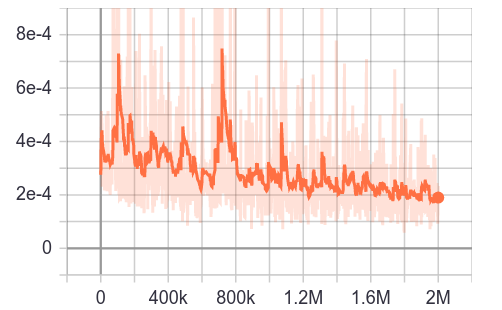}}
\newcommand{\linearval}{\includegraphics[height=0.5\linewidth]{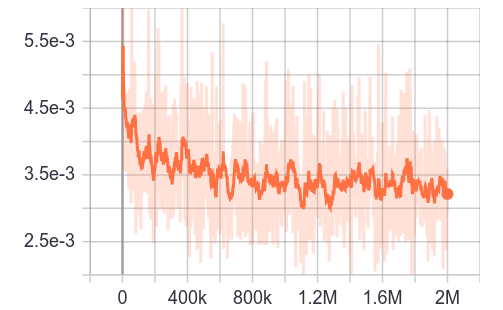}}

\begin{figure}
\setlength{\tabcolsep}{0pt}
\begin{center}
\resizebox{\linewidth}{!}{
\begin{tabular}{cc}
 \lineartrain & \linearval \\
 Training loss & Validation loss \\
\end{tabular}
}
\end{center}
\vspace{-.5em}
\caption{
Training and validation loss for \COTR with {\em linear positional encoding}. Both losses slowly converge to a stable status.
}
\vspace{-1em}
\label{fig:linear}
\end{figure}

\newcommand{\logtrain}{\includegraphics[height=0.5\linewidth]{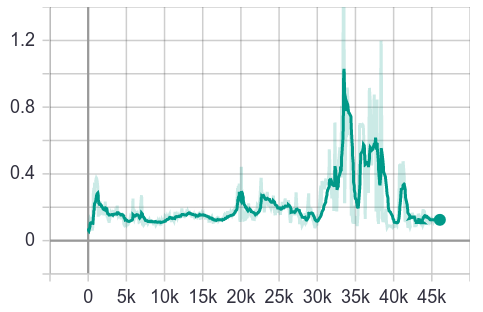}}
\newcommand{\logval}{\includegraphics[height=0.5\linewidth]{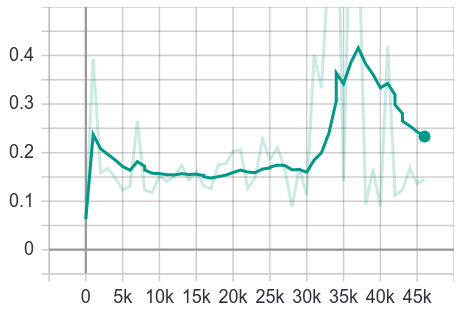}}

\begin{figure}
\setlength{\tabcolsep}{0pt}
\begin{center}
\resizebox{\linewidth}{!}{
\begin{tabular}{cc}
 \logtrain & \logval \\
 Training loss & Validation loss \\
\end{tabular}
}
\end{center}
\vspace{-.5em}
\caption{
Unstable training and validation loss for \COTR with {\em log-linear positional encoding}. We terminate the training earlier as the loss diverges.
}
\vspace{-1em}
\label{fig:log}
\end{figure}

\section{Architectural details for \COTR}

\paragraph{Backbone} 
We use the lower layers of ResNet50~\cite{He16} as our CNN backbone.
We extract the feature map with 1024 channels after layer3, \ie, after the fourth downsampling step.
We then project the feature maps with 1024 channels with $1\times1$ convolution to 256 channels to reduce the amount of computation that happens within the transformers.

\paragraph{Transformers} 
We use 6 layers in both the transformer encoder and the decoder. 
Each encoder layer contains an 8-head self-attention module, and each decoder layer contains an 8-head encoder-decoder attention module.
Note that we disallow the self-attention in the decoder, in order to maintain the independence between queries -- queries should not affect each other.

\paragraph{MLP} 
Once the transformer decoder process the results, we obtain a 256 dimensional vector that represents where the correspondence should be.
We use a 3-layer MLP to regress the corresponding point coordinates from the 256-dimensional latent vector.
Each layer contains 256 neurons, followed by ReLU activations.

\section{Architectural details for the MLP variant}

\paragraph{Backbone} 
We use the same backbone in \COTR. 
The difference here is that, once the feature map with 256 channels is obtained, we apply max pooling to extract the global latent vector for the image, as suggested in~\cite{Halimi20}.
We also tried a variant where we do not apply global pooling and use a fully-connected layer to bring it down to a manageable size of 1024 neurons but it quickly provided degenerate results, where all correspondence estimates were at the centre.

\paragraph{MLP} 
With the latent vectors from each image, we use a 3 layer MLP to regress the correspondence coordinates. 
Specifically, the input to the coordinate regressor is a 768-dimensional vector, which is the concatenation of two global latent vectors for the input images and the positional encoded query point.
Similarly to the MLP used in \COTR, each linear layer contains 256 neurons, and followed by ReLU activations.

\section{Comparing with \RAFTP}

\begin{table}
\setlength{\tabcolsep}{6pt}
\begin{center}
\resizebox{\linewidth}{!}{

\begin{tabular}{lcccccccc}
\toprule
\multirow{2}{*}{Method} & \multicolumn{8}{c}{ETH3D}                                                                                                                                                                                                              \\
\cmidrule(r){2-9}
                        & AEPE$\downarrow$ & rate=3 & rate=5 & rate=7 & rate=9 & rate=11 & rate=13 & rate=15 \\
\RAFT                    &                          & 1.92                       & 2.12                       & 2.33                       & 2.58                       & 3.90                        & 8.63                        & 13.74                       \\
\COTRnobf                    &                          & \textbf{1.66}              & \textbf{1.82}              & \textbf{1.97}              & \textbf{2.13}              & \textbf{2.27}               & \textbf{2.41}               & \textbf{2.61}               \\
\COTRTnobf           &                          & \underline{1.71}                       & \underline{1.92}                       & \underline{2.16}                       & \underline{2.47}                       & \underline{2.85}                        & \underline{3.23}                        & \underline{3.76}  \\
\bottomrule
\end{tabular}
} %

\resizebox{\linewidth}{!}{
\begin{tabular}{lcccccccc}
\multirow{2}{*}{Method} & \multicolumn{2}{c}{KITTI 2012}                    & \multicolumn{2}{c}{KITTI 2015}                    & \multicolumn{4}{c}{HPatches}                                     \\
\cmidrule(r){2-3}
\cmidrule(r){4-5}
\cmidrule(r){6-9}
                        & AEPE$\downarrow$ & Fl$\downarrow$ & AEPE$\downarrow$ & Fl$\downarrow$ & AEPE$\downarrow$          & PCK-1px$\uparrow$           & PCK-3px$\uparrow$           & PCK-5px$\uparrow$           \\
\RAFT                    & \underline{2.15}                     & \underline{9.30}                   & \underline{5.00}                     & 17.4                   & 44.3          & 31.22          & 62.48          & 70.85          \\
\COTRnobf                    & \textbf{1.28}            & \textbf{7.36}          & \textbf{2.62}            & \textbf{9.92}          & \textbf{7.75} & \textbf{40.91} & \textbf{82.37} & \textbf{91.10} \\
\COTRTnobf           & 2.26                     & 10.50                  & 6.12                     & \underline{16.90}                  & \underline{7.98}          & \underline{33.08}          & \underline{77.09}          & \underline{86.33}    \\     
\bottomrule
\end{tabular}
} %

\resizebox{\linewidth}{!}{
\setlength{\tabcolsep}{12pt}
\begin{tabular}{lccc}
\multicolumn{1}{c}{\multirow{2}{*}{Method}} & \multicolumn{3}{c}{Image Matching Challenge} \\
\cmidrule(r){2-4}
\multicolumn{1}{c}{} & Num. Inl.$\uparrow$ & mAA(5$^\circ$)$\uparrow$ &  mAA(10$^\circ$)$\uparrow$ \\
\midrule
\RAFT+DEGENSAC (N= 2048) & 1066.1                        & 0.163                      & 0.259                       \\
\COTRnobf+DEGENSAC (N= 2048) & \textbf{1686.2}               & \textbf{0.515}             & \textbf{0.678}\\
\bottomrule
\end{tabular}
} %

\end{center}
\vspace{-.5em}
\caption{
RAFT on ETH3D, KITTI, HPatches, and IMC2020.
}
\vspace{-1em}
\label{tbl:raft}
\end{table}

\RAFTP performs better in KITTI-type of scenarios, not necessarily so for other cases.
To show this, we provide results for \RAFTP on all other datasets in 
\Table{raft}.
On KITTI, sparse \COTRnobf still performs best, and with the interpolation strategy it is roughly on par with \RAFTP. %
On other datasets, \COTRnobf outperforms \RAFTP by a large margin\footnote{Note that \RAFTP requires two input images of the same size. We resize them to 1024$\times$1024 for HPatches and the Image Matching Challenge.}.

\end{document}